\documentclass[a4paper,conference]{IEEEtran}

\usepackage{booktabs}
\usepackage{amsfonts}

\ifCLASSOPTIONcompsoc
\usepackage[nocompress]{cite}
\else
\usepackage{cite}
\fi
\ifCLASSINFOpdf
\usepackage[pdftex]{graphicx}
\graphicspath{{./pdf/}{./jpeg/}}
\DeclareGraphicsExtensions{.pdf,.jpeg,.png,.jpg}
\else
\usepackage[dvips]{graphicx}
\graphicspath{{eps/}}
\DeclareGraphicsExtensions{.eps}
\fi
\ifCLASSOPTIONcompsoc
\usepackage[caption=false,font=normalsize,labelfont=sf,textfont=sf]{subfig}
\else
\usepackage[caption=false,font=footnotesize]{subfig}
\fi

\usepackage{fixltx2e}

\usepackage{stfloats}

\usepackage{url}
\begin{document}
	\title{A Gated and Bifurcated Stacked U-Net Module for Document Image Dewarping}
	
	\author{
		\IEEEauthorblockN{Hmrishav Bandyopadhyay}
		\IEEEauthorblockA{Dept. of Electronics and Telecomm. Engg.\\
			Jadavpur University, West Bengal, India\\
			hmrishavbandyopadhyay@gmail.com}
		
		\and 
		
		\IEEEauthorblockN{Tanmoy Dasgupta\IEEEauthorrefmark{1},
			Nibaran Das\IEEEauthorrefmark{2}, and
			Mita Nasipuri\IEEEauthorrefmark{3}}
		\IEEEauthorblockA{Dept. of Computer Science and Engg.\\
			Jadavpur University, West Bengal, India\\
			\IEEEauthorrefmark{1}tdg@ieee.org, \IEEEauthorrefmark{2}nibaran.das@jadavpuruniversity.in and\\ \IEEEauthorrefmark{3}mitanasipuri@gmail.com}
	}

	\maketitle
	
	\begin{abstract}
		Capturing images of documents is one of the easiest and most used methods of recording them. These images however, being captured with the help of handheld devices, often lead to undesirable distortions that are hard to remove. We propose a supervised Gated and Bifurcated Stacked U-Net module to predict a dewarping grid and create a distortion free image from the input. While the network is trained on synthetically warped document images, results are calculated on the basis of real world images. The novelty in our methods exists not only in a bifurcation of the U-Net to help eliminate the intermingling of the grid coordinates, but also in the use of a gated network which adds boundary and other minute line level details to the model. The end-to-end pipeline proposed by us achieves state-of-the-art performance on the DocUNet dataset after being trained on just 8 percent of the data used in previous methods.\\
	\end{abstract}
	

	\IEEEpeerreviewmaketitle

	\section{Introduction}
	With the rising popularity of smartphones and portable cameras, digitized documents have become a part and parcel of the common man's life. Document images, although captured easily with the help of these cameras, often lose the ability to represent the actual document due to inconsistencies in structure, illumination, and camera angles. The high variance of these document images from a flatbed-scanned version of the same make them unsuitable for further information extraction and analysis.

	Previous methods of dewarping with classical image processing techniques performed well when faced with standard curves or spherical distortions, but failed  when dealing with sharp folds and corner-type distortions. To rectify document images distorted with simultaneous folds and curves, various deep learning architectures have been proposed \cite{Ma-CVPR18,das2019dewarpnet}.

	We propose a document image dewarping model which outstrips the previous models in both speed and dewarp quality. The specific contributions of our work can be summarized as:
	
	\begin{figure}
		\centering
		\subfloat[Fold warp]{	
			\includegraphics[width=30mm]{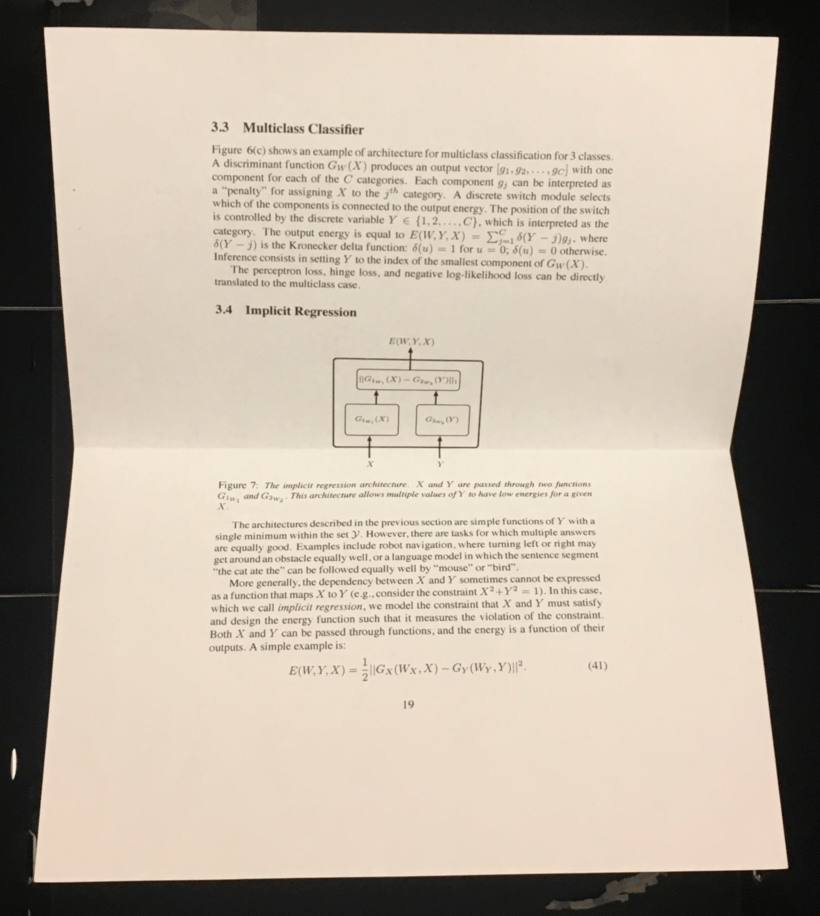}
		}
		\subfloat[Curve warp]{
			\includegraphics[width=30mm]{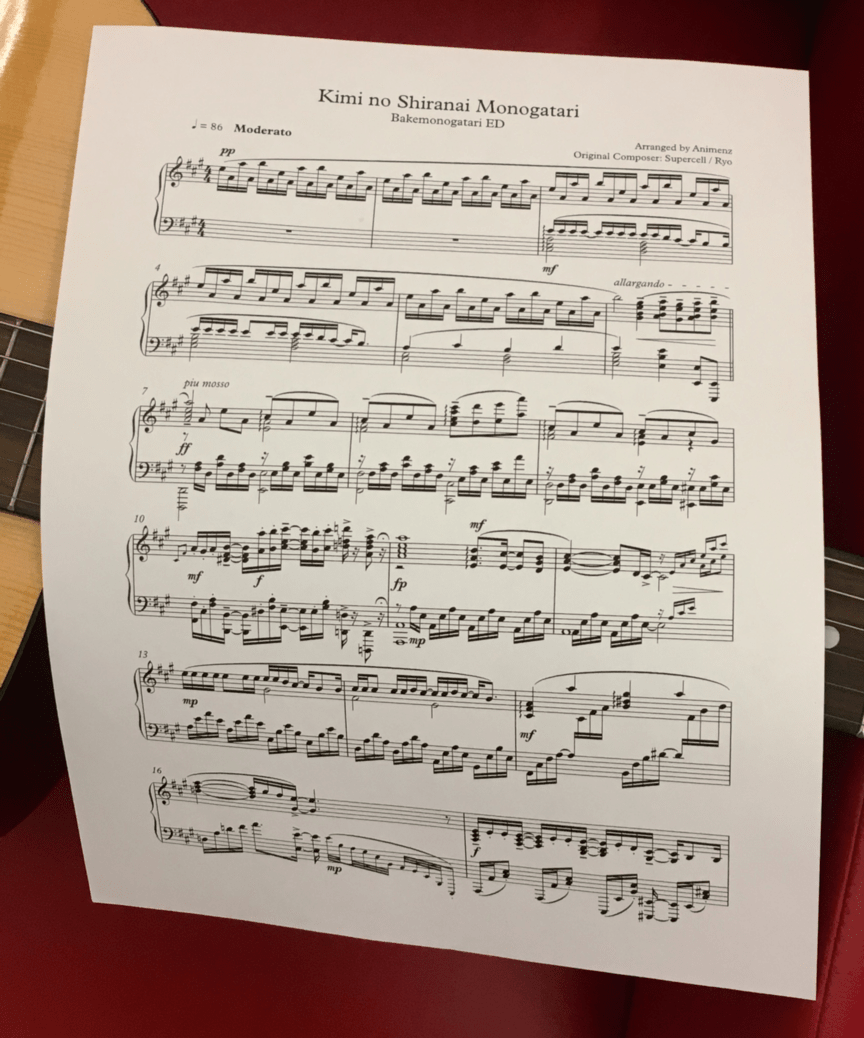}
		}
		
		\subfloat[Rectified Fold]{
			\includegraphics[width=30mm]{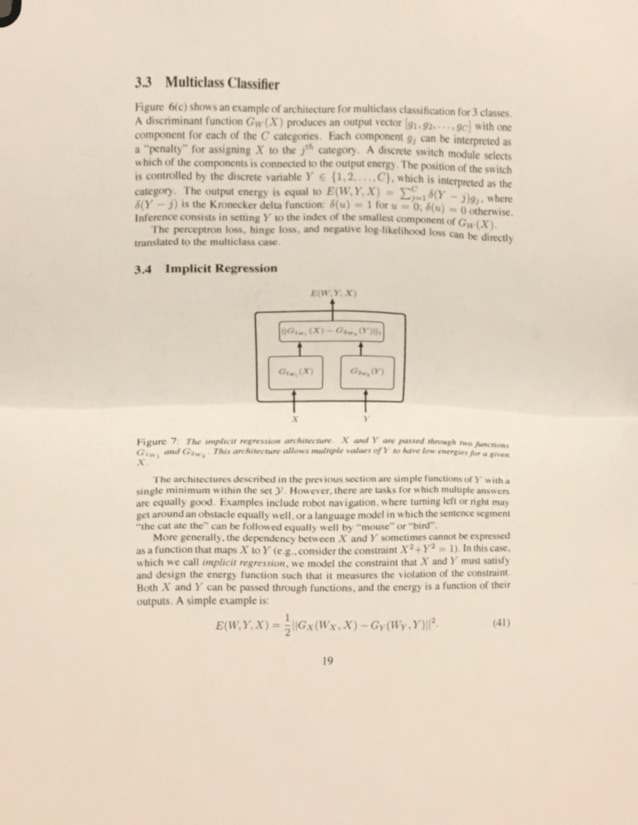}
		}
		\subfloat[Rectified Curve]{
			\includegraphics[width=30mm]{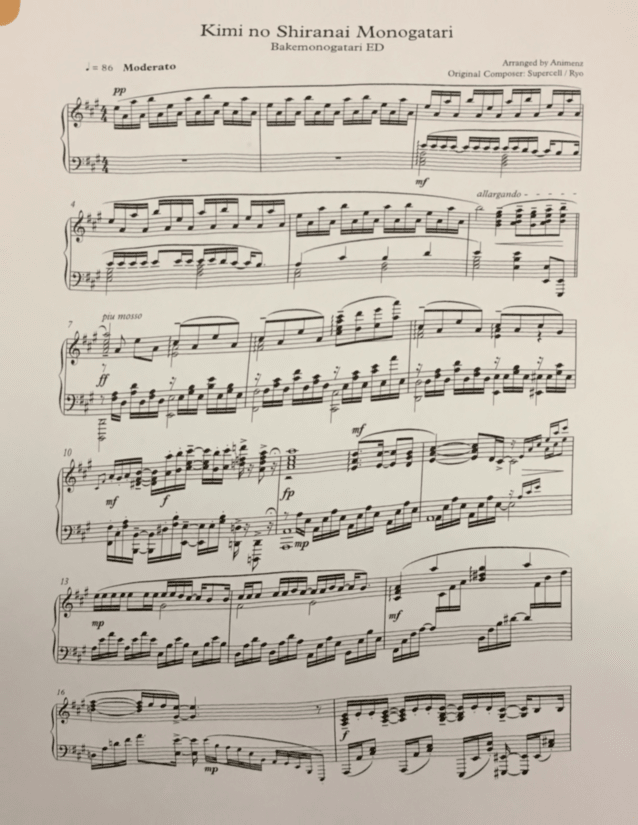}
		}
		\caption{Rectified Images on the basis of Inputs supplied to the end-to-end network}
	\end{figure}
	
	\begin{enumerate}
		\item Our network takes in 256x256 images as inputs to produce an unwarping grid which can be interpolated to reconstruct images at their original resolution. The parameters are learned efficiently and as such the model learns in just 8 thousand images, \emph{less than one-tenth of the dataset sizes in previous end-to-end methods}.
		
		\item We propose a bifurcated U-Net as the secondary U-Net of our stacked U-Net system to  help in channel level segregation while predicting dense grid unwarps.
		
		\item A gated branch of the primary U-Net is proposed following \cite{GSCNN} that	enables the secondary U-Net to recognize lines and boundaries in the warped document image.
		
	\end{enumerate}

	\section{Previous work}
	
	Previous literature has studied the problem of document unwarping through both single and multi image tasks. Vision systems have been designed that make use of well calibrated stereo cameras or structured light projectors ~\cite{1334171,Ulges,937649,meng2014active} to get insight into distortion factors present within the document. Specifically, Meng \textit{et. al.}\cite{meng2014active} set up a platform with structured laser beams for acquiring curls in the warped document images. Although these systems give high quality results, their application is severely bound due to the limitation posed by additional hardware.

	Tsoi and Brown \cite{tsoi2007multi} came up with a way to  reduce hardware by utilizing the boundary information from multi-view images and generated the rectified image. Multi view systems like these which require more than one image for reconstructing 3-D shapes can do without additional hardware, but would always require more than one image, which essentially serves as another limitation.

	Dewarping methods involving single images although free from these limitations, pose a bigger problem as the two dimensional image is now used to get a three dimensional perspective of the document. Here, we find the use of not only classical image processing and machine learning methods, but also deep learning models.

	Techniques involving classical image processing have been able to sufficiently dewarp linearly warped images of documents\cite{wu2002document, lu2006document, mollah2009fast}, but have been forced into a corner when non-linearly warped or folded-document images have been tested on them\cite{shafait2007document}. While some of them have passed with considerably good values in one evaluation metric, they have been severely penalized in another. Ezaki \textit{et. al.} \cite{ezaki2005dewarping} proposed a global optimization based dewarping technique concentrated on converting warped lines to parallel ones and thus, rectifying non-linearly warped document images. Similar work involving curled text-line detection based techniques were demonstrated in \cite{ulges2005document, kakumanu2006document}. Boundary fitting techniques like \cite{wu2007model, wu2008model, masalovitch2007usage, he2013book, huggett2013method, bolelli2017indexing} , however, proved to be considerably more effective in solving this problem when compared with techniques involving detection and rectification on a line level.

	An approach involving segmentation of text lines for detection of warps was proposed by Gatos \textit{et. al.} \cite{gatos2007segmentation}. Similar works involved the warped text-line being represented as texture restricted by two smooth curved lines on the top and bottom and were described in \cite{chethan2010image,kwon2016method}. These methods however, involve techniques which rely far too much on the text to image ratio of the document. Typically, such segmentation methods were found to fail in documents with a larger proportion of images.

	We find better text-line detection in \cite{frinken2011novel, koo2016text} which can handle fairly complex page layouts. They take up iterative checks for alignment of text lines in images and make repairs wherever needed. These techniques, although being highly successful in evaluation metrics, are iterative and thus inherently slow. Moreover, their heavy reliance on the availability of clear boundaries make them impractical in real world scenarios.
	
	CNN based methods for document dewarping were employed by Das \textit{et. al.} in \cite{das2017common} where the application was limited to detection of paper creases for rectification. The first end to end deep learning model for dewarping of document images was proposed by Ma \textit{et. al.} in \cite{Ma-CVPR18}. Along with the development of an end to end model, \cite{Ma-CVPR18} also made available for the first time, a method of generating huge warped document datasets. Recently, Das \textit{et. al.} came up with a way of generating more realistic document images with illumination effects in \cite{das2019dewarpnet}.
	
	\section{Dataset}
	The recent availability of synthetic datasets has resulted in a sudden spur in the use of deep learning in this domain. We generate a synthetic dataset as proposed by \cite{Ma-CVPR18}. The task of generation of the dataset proceeds by first generating a perturbed mesh. This mesh then provides the sparse deformation field needed to build a dense warping map. This warping map when applied to the original image can generate a distorted image of the document. With the help of random warping maps , we generate 8K images consisting of folds and curves.
	To incorporate realistic effects on the generated data, we make use of texture images available in Describable Texture Dataset (DTD) \cite{cimpoi2014describing} and apply background textures to generated warped images. 
	
	\section{Approach}
	
	\begin{figure}
		\centering
		{	
			\includegraphics[width=70mm]{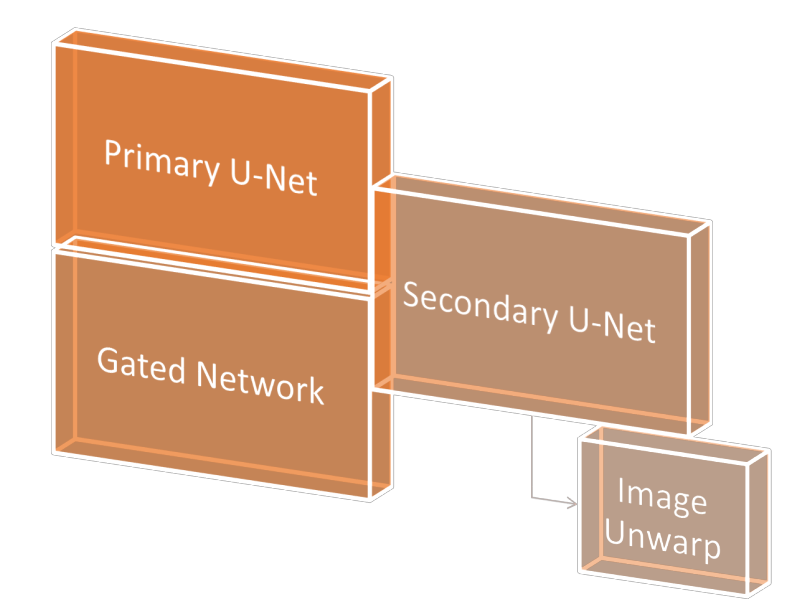}
		}
		\caption{Complete Architecture}
		\label{fig:arch}
	\end{figure}
	
	Our Network as in Fig. \ref{fig:arch} makes significance changes to a stacked U-Net backbone in order to help recognize warpings in document images and to dewarp them. Specifically,  we propose the use of a gated network for identification of lines and boundaries and pass them onto the second U-Net concatenated with data received from the first U-Net. We also make use of a bifurcation in the last decoder to get us fairly independent channel values for the dense grid prediction.

	The network takes deformed document images as input $I \in \mathbb{R} ^{h\times w\times3}$ and generates a dense grid prediction $G \in \mathbb{R} ^{hxwx2}$. With this grid, an unwarp of the input image can be performed to get the distortion free document. Broadly, the network can be divided into the primary and secondary stacks of the U-Net.
	
	\subsection{The Primary U-Net}
	
	\begin{figure}[!t]
		\includegraphics[width=90mm]{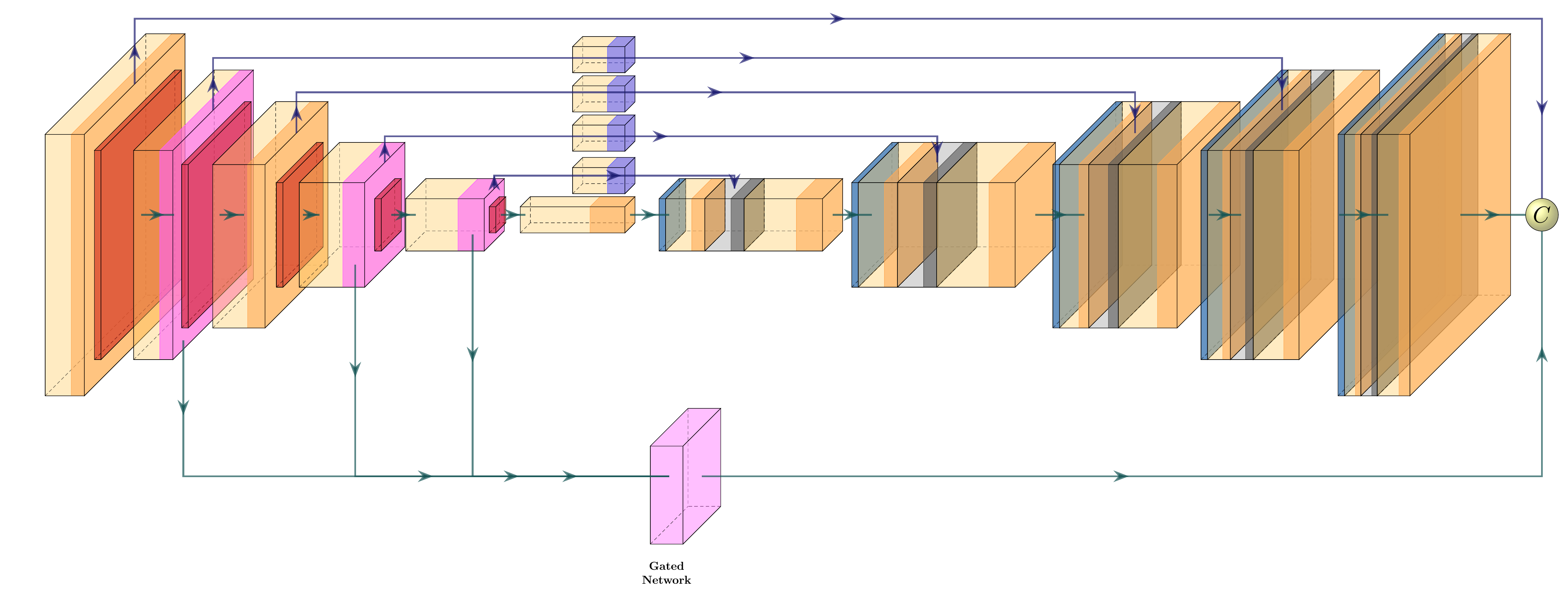}
		\caption{Primary U-Net}
		
		\label{fig:arch1}
	\end{figure}
	
	The primary U-Net as in Fig. \ref{fig:arch1} or the first of the stacked U-Net series contains a set of up-sampling and down-sampling layers. The main deviation from a traditional U-Net can be seen in the presence of a gated network.
	\\

	\begin{figure}
		\centering
		{	
			\includegraphics[width=70mm]{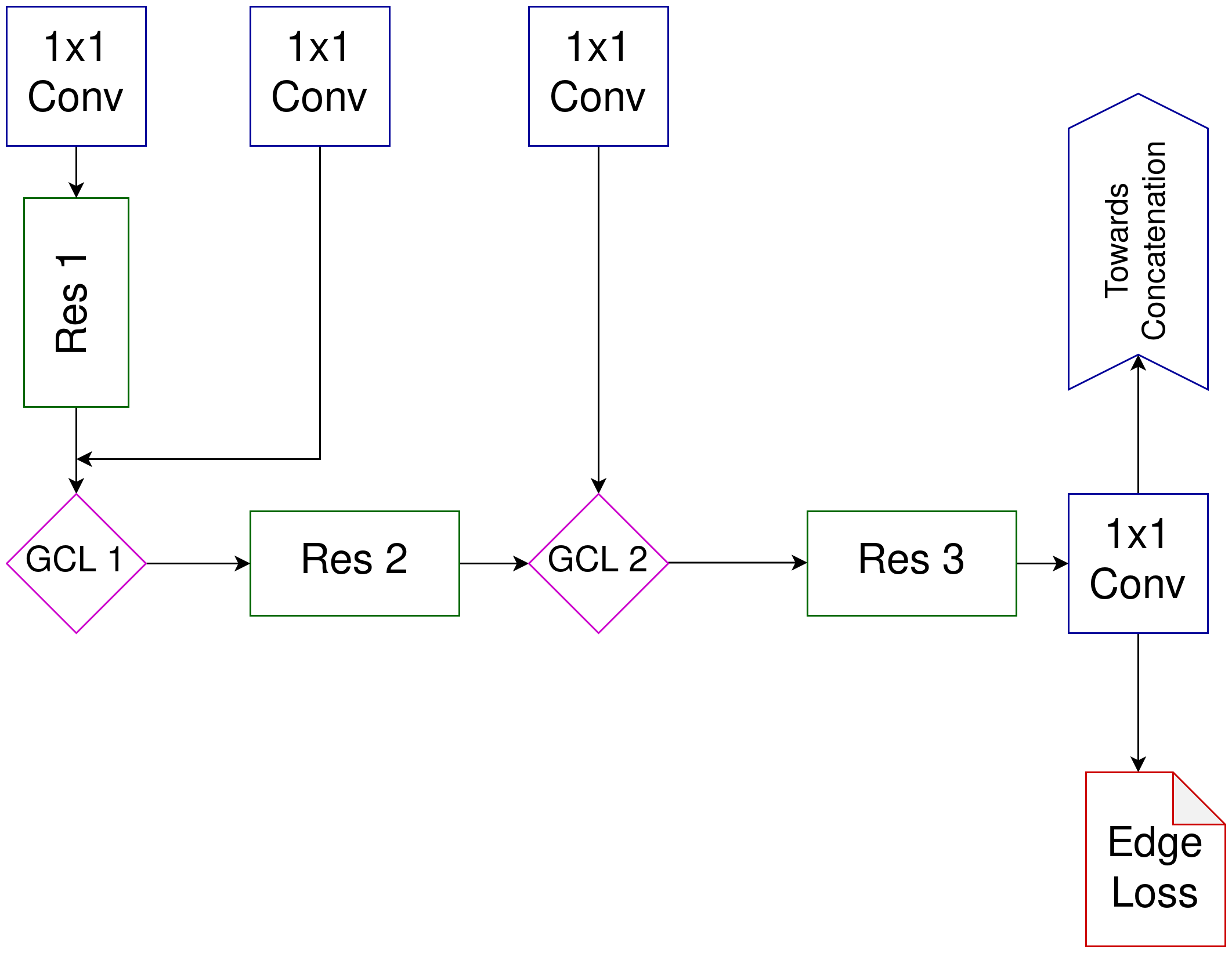}
		}
		\caption{Gated Convolutional Network}
		\label{fig:arch3}
	\end{figure}
	The Gated Network [Fig. \ref{fig:arch3}] extracts data from layers before the 2nd, 4th and 5th poolings. It primarily consists of Gated Convolutional Layers (GCLs) as proposed in \cite{GSCNN}. The presence of gates in the layers helps the model reject information not related to edges or boundaries. As a result of this, we have layers that deal only with edge data. The GCL works by forming attention maps through the layers extracted and formulates convolutions based on them. Both attention map computations and GCLs being differentiable, backpropagation can be performed end-to-end.
	The edge level detail passed by the gated network serves to `inform' the CNN of boundaries and orientations.
	
	The skip connections of a traditional	 U-Net in the U-Net modules are replaced by a CNN path with ReLU non-linearity to help in data extraction at different spatial levels before being fed to the concurrent decoders.
	\\
	
	The primary U-Net module thus takes in the image of the deformed document $I \in \mathbb{R} ^{256\times256\times3}$ and passes it through a series of encoders to get the bottleneck of $B \in \mathbb{R} ^{1024\times8\times8}$. The layers $L_{2} \in \mathbb{R} ^{64\times128\times128},L_{4}\in \mathbb{R} ^{256\times32\times32},L_{5}\in \mathbb{R} ^{512\times16\times16}$ are extracted and go through the gated module $G$. Finally, the decoded outputs $O\in \mathbb{R} ^{2\times256\times256}$, gated network outputs $G_{o}\in \mathbb{R},  ^{16\times256\times256}$ and the initial convolution outputs $X\in \mathbb{R} ^{32\times256\times256}$ are concatenated to produce the output of the primary U-Net $U_{1}\in \mathbb{R} ^{50\times256\times256}$. 
	
	\subsection{The Secondary Bifurcated U-Net}
	\begin{figure*}[!h]
		\centering
		{	
			\includegraphics[width=0.7\linewidth]{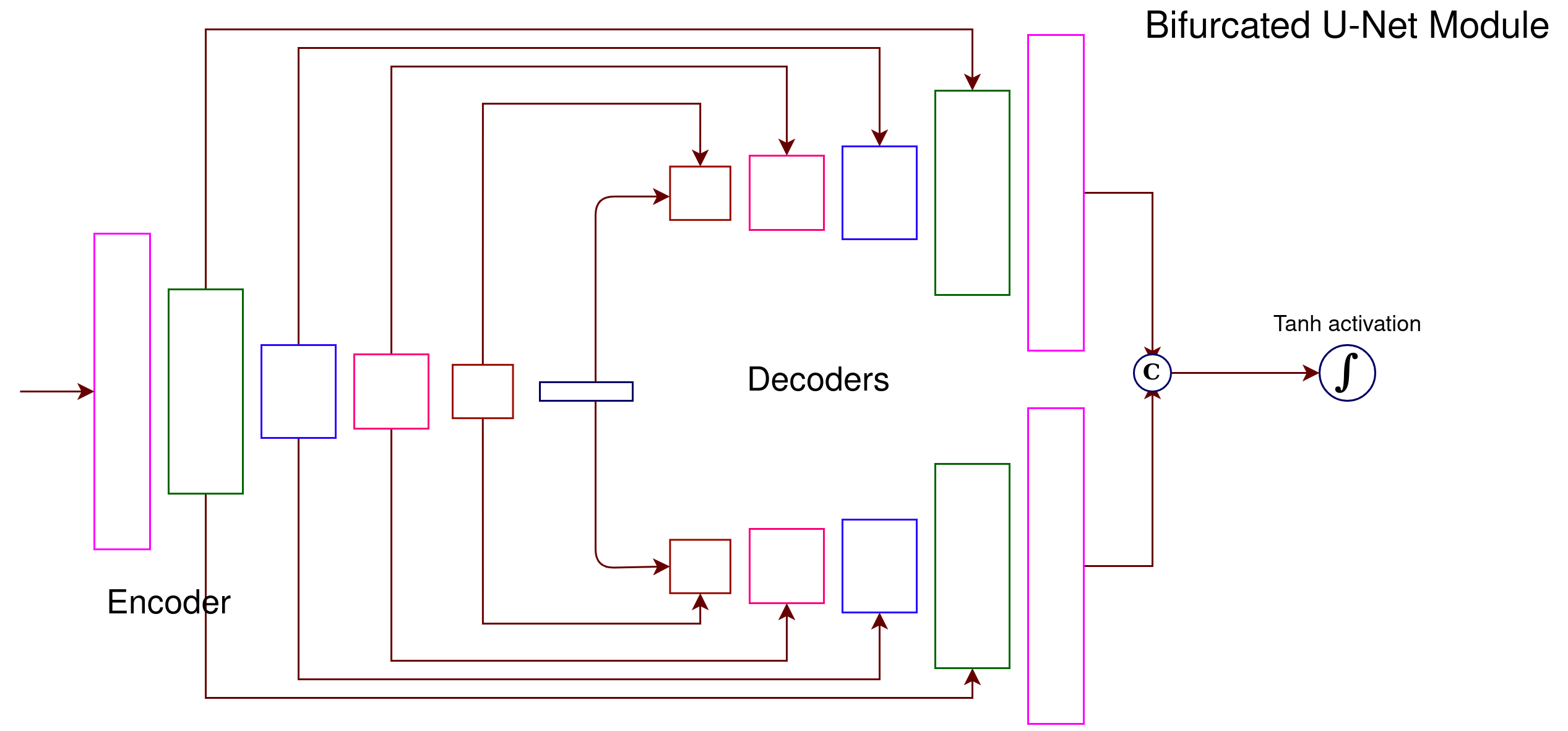}
		}
		\caption{Secondary U-Net}
		\label{arch4}
	\end{figure*}
	
	Convolutional networks fail spectacularly when dealing with coordinate data as shown by \cite{coordconv}.To solve this while making grid predictions, \cite{das2019dewarpnet} uses the modules suggested in \cite{coordconv}. However, we find that a more task specific network can be designed which can enhance the ability of CNNs while working with a dense grid prediction.
	
	The general CNN works by summing computed data across all input channels for specific window sizes. Ultimately the number of channels in the output is the number of filters that the convolutional block contains. We came to the conclusion that using a single decoder in the final U-Net block would mean that although 	information is extracted in all blocks, \emph{only} the last two convolutional filters would decode the grid values into their respective \emph{channels} for the final output. To get round this issue, we came up with the usage of multiple decoder blocks for the single secondary U-Net encoder. 
	\\
	
	From our experiments, we found a spike in the results as soon as we shifted from a single decoder to multiple decoders with shared weights. This further strengthened our case as the decoder with shared weights, although restricted by weight sharing, had the advantage of being decoded separately right from the bottleneck layer. Finally, we found the best results when we lifted the restriction of weight sharing and made the decoders independent of each other. In our network, the encoder output is split into two and is sent separately to the two decoders. The output from these decoders is concatenated and normalized with the Tanh activation function to provide us the output of the network.
	\\
	
	The output of the primary U-Net $U_{1}\in \mathbb{R} ^{50\times256\times256}$ is fed as input to the secondary U-Net. The Bottleneck $B\in \mathbb{R} ^{1024\times8\times8}$ is split into $B_{1}\in \mathbb{R} ^{512\times8\times8}$ and $B_{2}\in \mathbb{R} ^{512\times8\times8}$. These blocks go through the decoders to give outputs $O_{1}\in \mathbb{R} ^{256\times256\times1}$ and $O_{2}\in \mathbb{R} ^{256\times256\times1}$, which are concatenated and normalized by a Tanh activation function to get the final grid $g\in \mathbb{R} ^{256\times256\times2}$.

	\section{Loss Functions}
	
	The loss function that helps train our network can be expressed as a summation of an edge loss and a grid loss, weighted with an appropriate $\lambda$.
	
	\subsection{Edge Loss}
	The edge loss can be expressed as a binary classification loss as we try to predict if a pixel belongs to the `edge class' or not. We make use of the Binary Cross Entropy loss function for this. Comparison of the output of the gated network is done by taking an edge detection function as the ground truth. The edge loss can be expressed in mathematical terms as:
	\[
	\mathcal{L}_{e}=-\frac{1}{N}\sum_{i=0}^{N}y_i.log(\hat{y_{i}})+(1-y_{i}).log(1-\hat{y_{i}})
	\]
	Where $\hat{y_{i}}$ represents pixel wise value of the predicted output and $y_{i}$ gives the ground truth measure.
	\subsection{Grid Loss}
	The grid loss trains not only the secondary U-Net but also the primary one as the loss is propagated throughout the model. Although previous methods as \cite{Ma-CVPR18} suggest the use of Mean Absolute Error (L1) to train the network against the ground truth grid, from our experiments we find the Least Squares method (L2) to be more robust. In fact, using Least Squares loss also helps our training algorithm to converge to a minima in a smoother way as compared to using L1 loss.
	Mathematically, we express grid loss as:
	\[
	\mathcal{L}_{g}=\frac{1}{N}\sum_{i=0}^{N}(g_{i}-\hat{g_{i}})^2
	\]
	\\
	\\
	Expressing the combined loss function, we have
	
	\[
	\mathcal{L}=\frac{1}{N}\sum_{i=0}^{N}(g_{i}-\hat{g_{i}})^{2}-\lambda.\frac{1}{N}\sum_{i=0}^{N}y_i.log(\hat{y_{i}})+(1-y_{i}).log(1-\hat{y_{i}})
	\]
	We take the value of $\lambda$ as 0.9 for all our experiments.
	
	\begin{figure}
		\centering
		\subfloat[Input Image]{	
			\includegraphics[width=30mm]{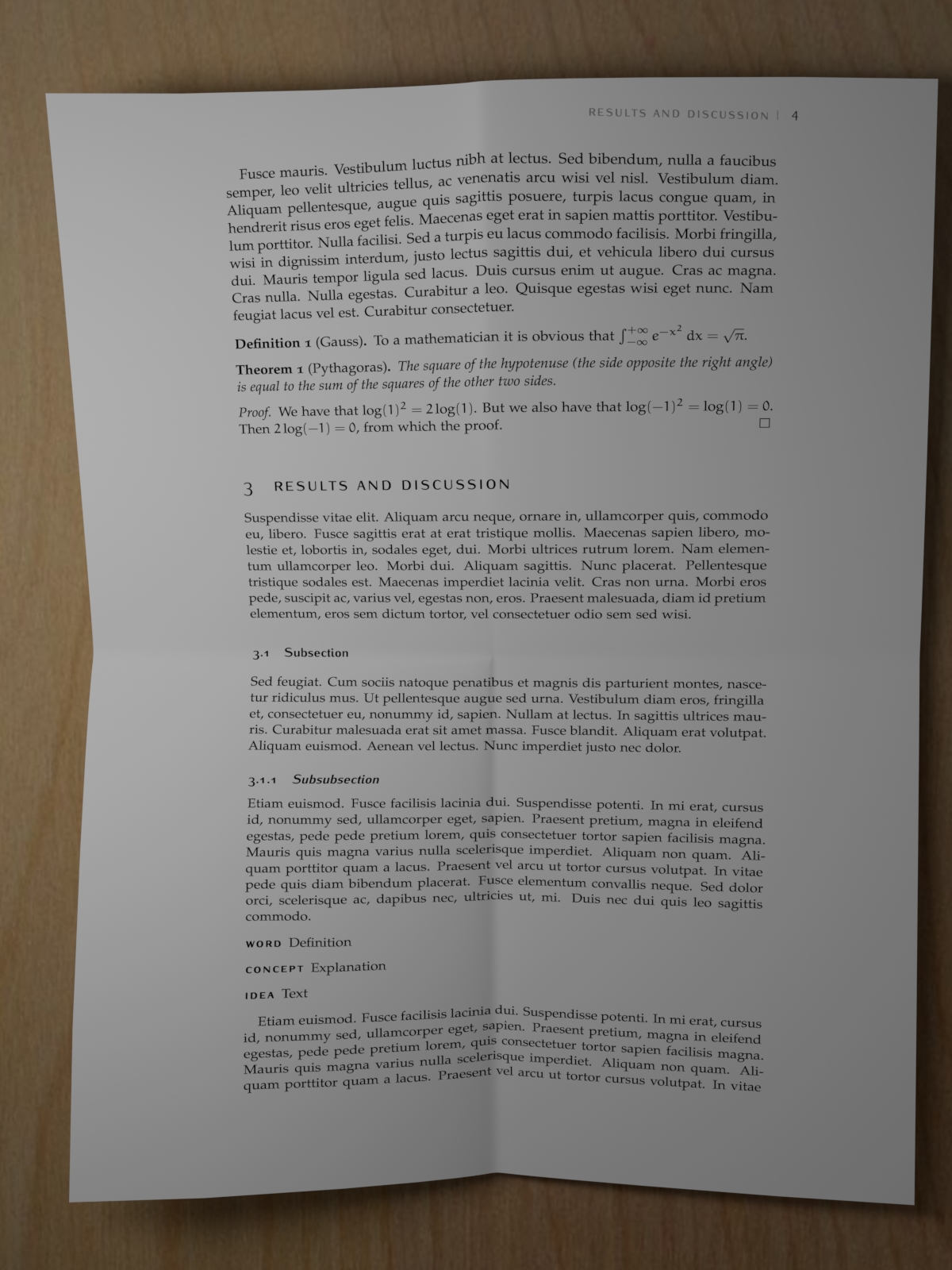}
		}
		\subfloat[Results from Das \textit{et. al.}]{
			\includegraphics[width=30mm]{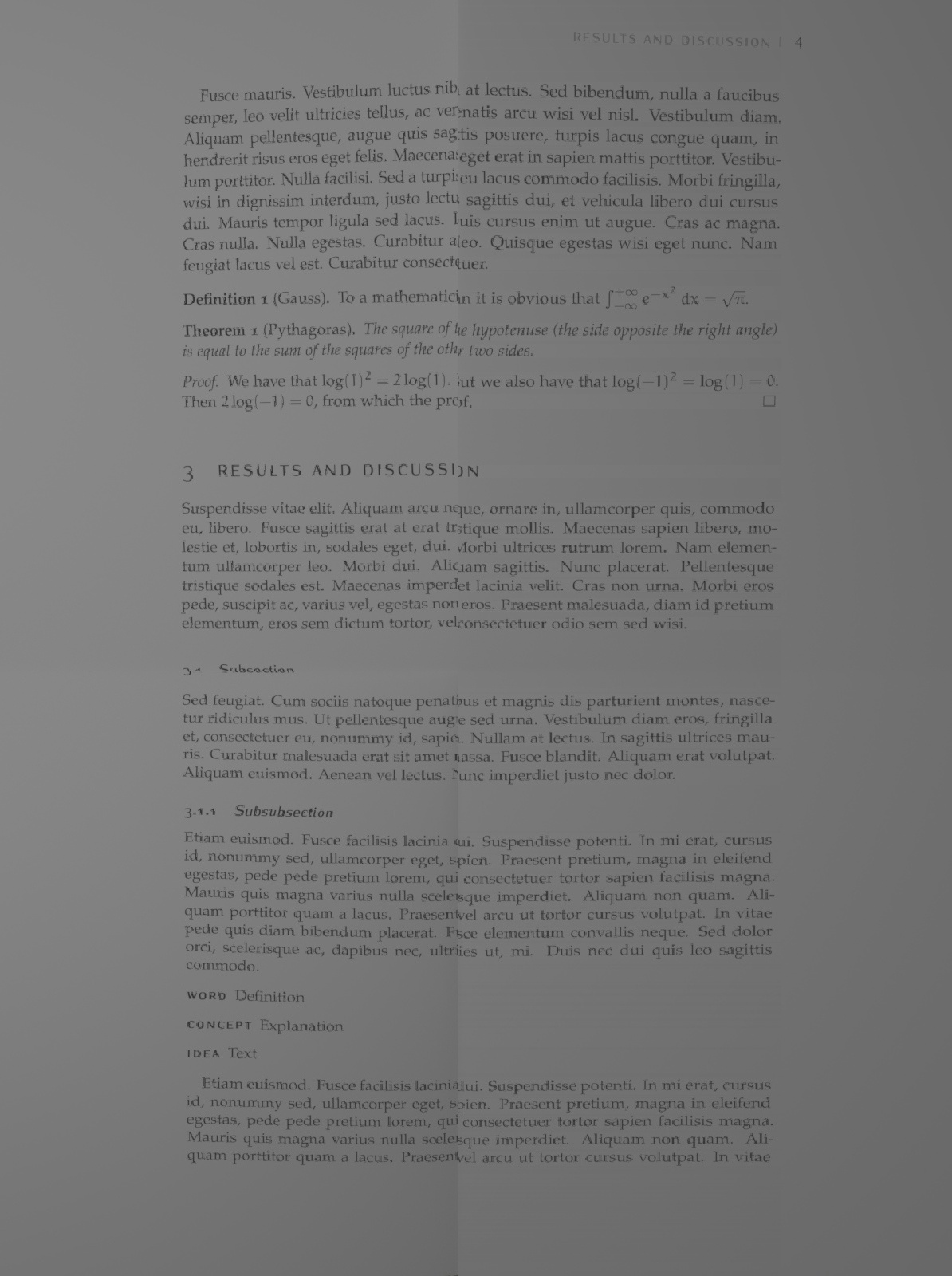}
		}
		
		\subfloat[Image edges from gated network]{
			\includegraphics[width=30mm]{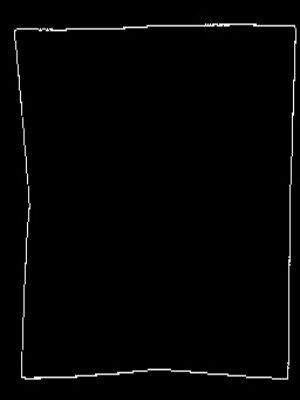}
		}\subfloat[Results from end-to-end model]{
			\includegraphics[width=30mm]{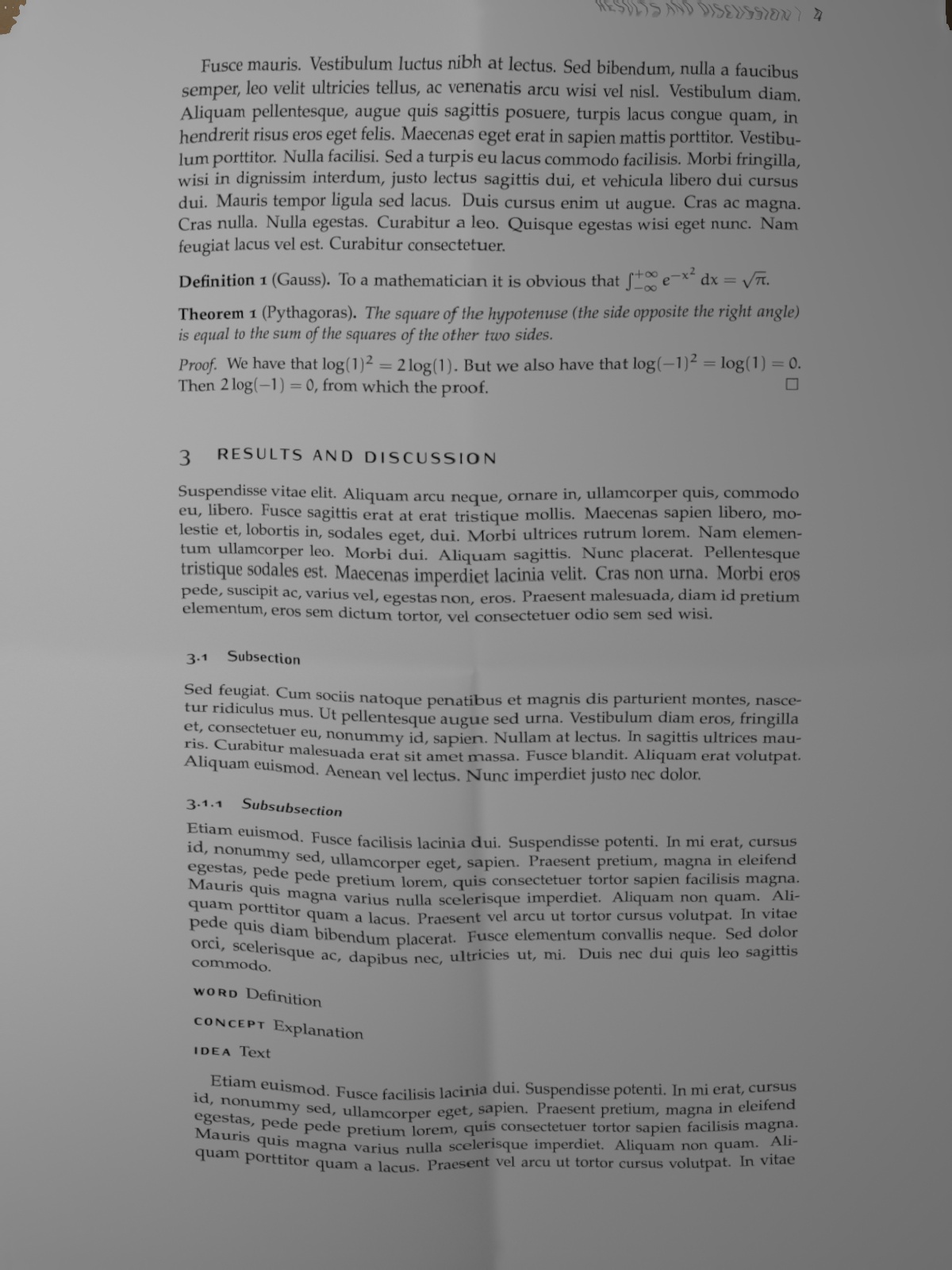}
		}
		\caption{Comparison with Das \textit{et. al.}\cite{das2017common}}
	\end{figure}
	
	\section{Evaluation Scheme}
	
	\begin{figure*}
	\centering
	\subfloat{	
		\includegraphics[width=30mm,height=40mm]{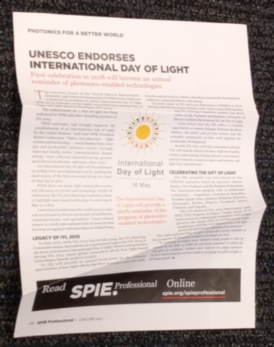}
	}\hspace*{-0.5em}\subfloat{
		\includegraphics[width=30mm,height=40mm]{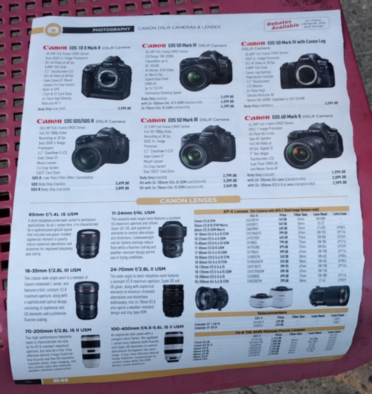}
	}\hspace*{-0.5em}\subfloat{
		\includegraphics[width=30mm,height=40mm]{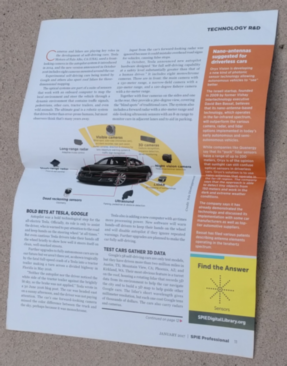}
	}\hspace*{-0.5em}\subfloat{
		\includegraphics[width=30mm,height=40mm]{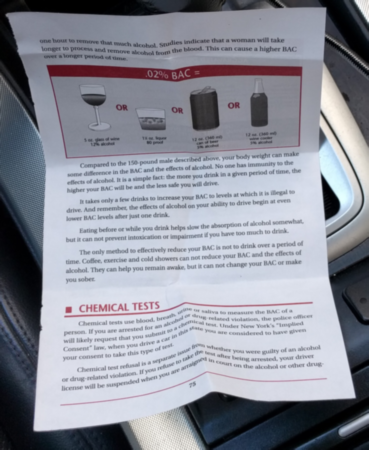}
	}
	\\[-0.7em]
	\subfloat{	
		\includegraphics[width=30mm,height=40mm]{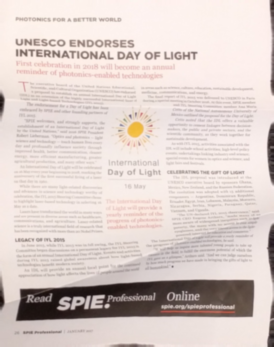}
	}\hspace*{-0.5em}\subfloat{
		\includegraphics[width=30mm,height=40mm]{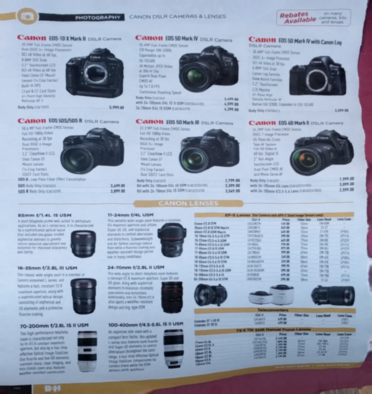}
	}\hspace*{-0.5em}\subfloat{
		\includegraphics[width=30mm,height=40mm]{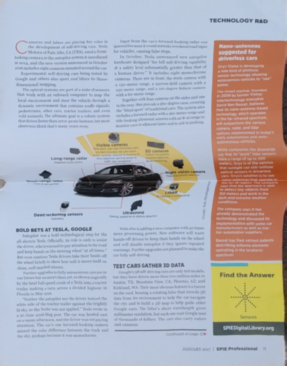}
	}\hspace*{-0.5em}\subfloat{
		\includegraphics[width=30mm,height=40mm]{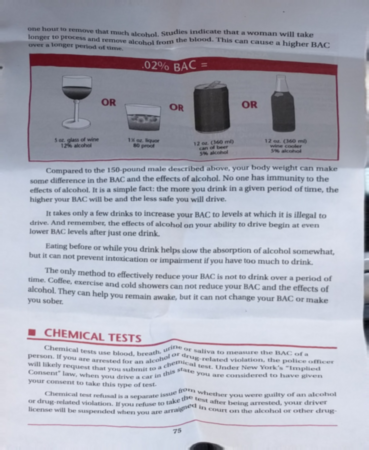}
	}\\[-0.7em]
	\subfloat{	
		\includegraphics[width=30mm,height=40mm]{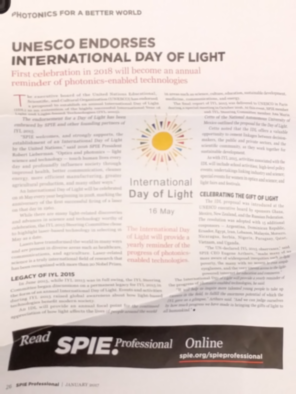}
	}\hspace*{-0.5em}\subfloat{
		\includegraphics[width=30mm,height=40mm]{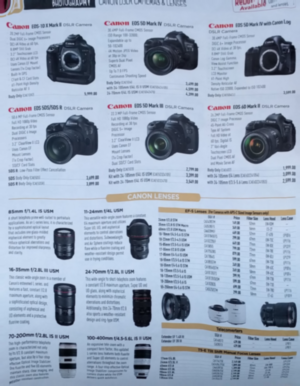}
	}\hspace*{-0.5em}\subfloat{
		\includegraphics[width=30mm,height=40mm]{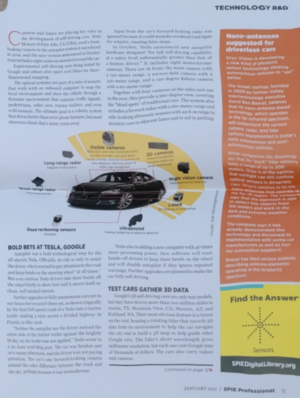}
	}\hspace*{-0.5em}\subfloat{
		\includegraphics[width=30mm,height=40mm]{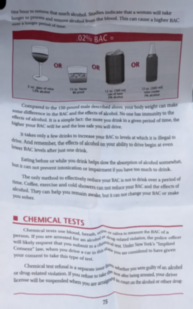}
	}\\[-0.7em]
\subfloat{	
	\includegraphics[width=30mm,height=40mm]{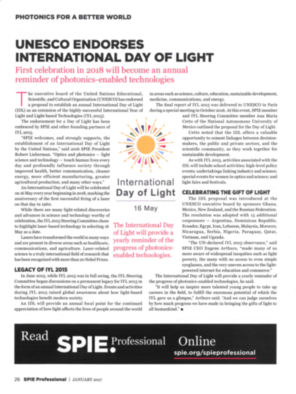}
}\hspace*{-0.5em}\subfloat{
	\includegraphics[width=30mm,height=40mm]{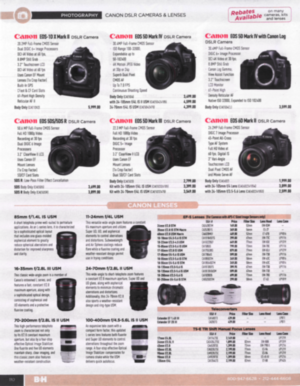}
}\hspace*{-0.5em}\subfloat{
	\includegraphics[width=30mm,height=40mm]{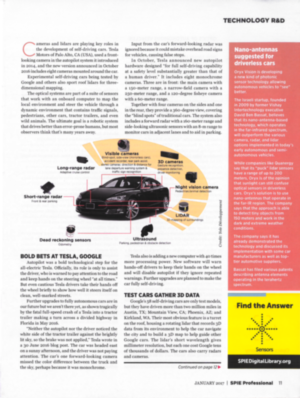}
}\hspace*{-0.5em}\subfloat{
	\includegraphics[width=30mm,height=40mm]{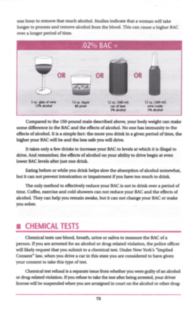}
}
	\caption{Rows from top to bottom: Cropped Images; DewarpNet Results; Our Results; Ground Truth}
\end{figure*}

	For the evaluation of our network and comparison with images of different sizes, previous methods suggest use of evaluation metrics like image similarity and Optical Character Recognition (OCR). OCR, however is highly dependent on the image to text ratio in the document and fails when the document comprises primarily of images. The limitation of application of this metric in all images is why we choose image similarity as the evaluation metric for our methods.

	In order to compare the similarity of output images and the scanned ground truth, we make a comparative study of results from image similarity metrics like MS-SSIM (Multi-Scale Structural Similarity Index), LD (Local Distortion), and SSIM (Structural Similarity Index) at various pyramid levels of the images.Although MS-SSIM is essentially a weighted average of SSIM applied across levels, we take note of SSIM values in order to realize how the unwarping quality varies along different scales of the image.
	\\
	\\
	\begin{figure}
		\centering
		\includegraphics[width=80mm]{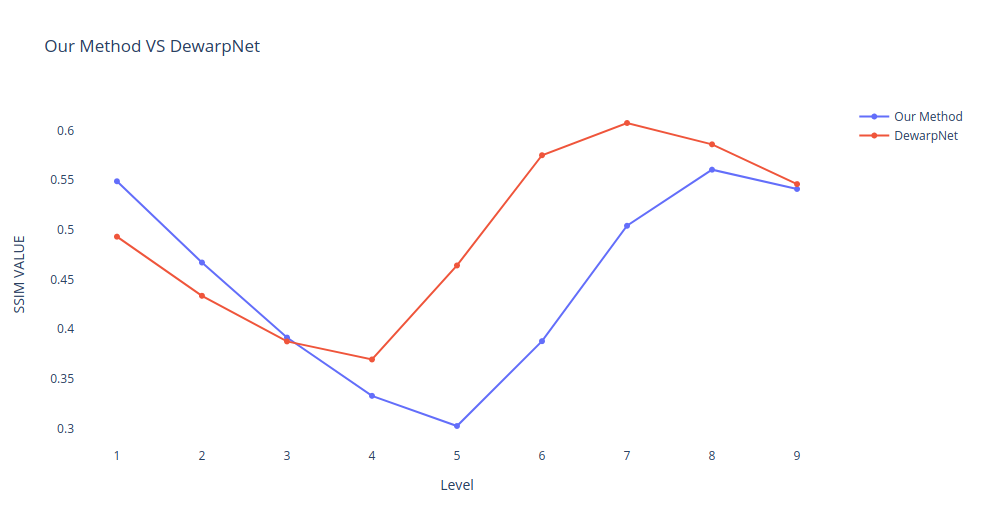}
		\caption{Comparison with DewarpNet Results. Our Model can be seen to have a higher SSIM till the 3rd level, after which, DewarpNet gets the upper hand. Level here refers to Gaussian Pyramid levels. The first level is the original image while the nth level has been down-sampled n-1 times}.
	\end{figure}
	\begin{table}[h]
	\centering 
	\caption{SSIM values on varying levels}
	\begin{tabular*}{\linewidth}{p{70pt}p{50pt}p{70pt}}    \toprule
		\textbf{Level} & \textbf{Our Method}&\textbf{DewarpNet \cite{das2019dewarpnet}}\\\midrule
		 \textbf{1} & \textbf{0.548915}& \textbf{0.493146} \\
		2& 0.467136&0.433653  \\
		3&0.39162&0.387747\\
		4&0.332977&0.369569\\
		5&0.302610&0.464170\\
		6&0.387984&0.575128\\
		7&0.504144&0.607561\\
		8&0.560574&0.586102\\
		9&0.541162&0.546075\\
		\hline
	\label{table:tab1}
	\end{tabular*}
\end{table}
	\section{Results}
	
	The MS-SSIM evaluations are performed on the DocUNet Image dataset \cite{Ma-CVPR18}. A 5-level-pyramid is used for calculating MS-SSIM where the weight for each	level is set at 0.0448, 0.2856, 0.3001, 0.2363, 0.1333 respectively following the original implementation \cite{wang2003multiscale}.
	
	The official implementation for the network discussed in here has been made open source and can be accessed at \emph{'https://github.com/DVLP-CMATERJU/RectiNet'}

	Although comparison has been made across DocUNet \cite{Ma-CVPR18}, DewarpNet\cite{das2019dewarpnet}, and our method, DewarpNet uses the Doc3D dataset while we use the DocUNet dataset for training. As due to unforeseen circumstances the Doc3D dataset has accessibility issues, we have been forced to restrict our experiments to the DocUNet dataset only. The fact that the Doc3D dataset is much better and more realistic than the DocUNet dataset, as seen in their tests, must be taken into consideration while comparing the results of both methods. 
	\\
	
	\emph{It should also be noted that the results in \cite{das2019dewarpnet} are on specific 880x680 area based resizes while we compare results on full resolution scanned images. This might account for any differences in our  results from their tabulated data. Furthermore, the results in DocUNet\cite{Ma-CVPR18} are results we obtained from their tabulations as their code was not made publicly available. As such, their results still represent values in 880x680 area patches and are for reference only.}
	\\

	We observe that our model achieves state-of-the art results in terms of SSIM values. It dominates the SSIM board till before the 4th down-sample, after which DewarpNet takes over. We would like to draw attention to the fact that the fourth down-sample would mean images have been reduced to $\frac{1}{256}$ of their original area and as such have been blurred almost completely. In-spite of leading in SSIM for 3 levels, we find our MS-SSIM metric lower than that of DewarpNet as DewarpNet's SSIM values for the 4th and 5th down-sample are much higher than ours in the previous three levels. Our model performs considerably better when compared with \cite{Ma-CVPR18} and \cite{tian2011rectification}.
	
	\begin{table}[h]
	\centering 
	\caption{MS-SSIM and LD values on comparison with other methods\label{tab:techniques}}
	\begin{tabular*}{\linewidth}{p{70pt}p{50pt}p{50pt}}    \toprule
		\textbf{Method} & \textbf{MS-SSIM} $\uparrow$ &\textbf{LD}$\downarrow$\\\midrule
		 Tian \textit{et. al.} \cite{tian2011rectification}&0.13* & 33.69\\
		 DocUNet \cite{Ma-CVPR18}&0.410*&14.08\\
		 \textbf{Our Method} &\textbf{0.415}&\textbf{13.2}\\
		 DewarpNet \cite{das2019dewarpnet}&0.437&8.98\\
		\hline
		
	\end{tabular*}
*: Results obtained from research reports where images have been scaled
\end{table}
	
	\section{Ablation Experiments}
	In order to test the effectiveness of our methods, we strip the model of the gated network and set up shared weights in the bifurcated U-Net . Aligning with our line of thought, we observe that the MS-SSIM values show a dip by \textbf{6.40\%} and \textbf{3.8 \%} when we remove the gated network and add the shared weights respectively. The exact values can be seen in Table \ref{table:tab3}. It should be noted that these values are lower than the reported values from \cite{Ma-CVPR18} although they have similar backbone because of the difference in our training dataset lengths.
	
	\begin{table}[h]
	\centering 
	\caption{MS-SSIM values on different features of network\label{tab:techniques}}
	\begin{tabular*}{\linewidth}{p{150pt}p{40pt}}    \toprule
		\textbf{Feature} & \textbf{MS-SSIM}\\\midrule
		  Shared Weights accross Decoder&0.399\\
		 Gated Network removal &0.388\\
		 \textbf{Actual Method} &\textbf{0.415}\\
		 
		\hline
	\label{table:tab3}
	\end{tabular*}

\end{table}

	\section{Conclusion and Future Work}
	In this paper, we presented a model capable of predicting dense grid mapping for dewarping 2D document images. We demonstrated the effectiveness of a bifurcated U-Net for dense grid predictions when compared to a traditional U-Net. We further enhanced our methods by the addition of a gated network for incorporating edge and line level details. The effectiveness of our model was exhibited when we reached DewarpNet level results while training our network on only 8K images.
	\\
	
	We find our results limited in certain respects. Comparing DewarpNet and our method, we observe that our method leaves out boundaries from the original document image while DewarpNet inculcates extra regions in the output. These strongly signify the need of a document localization module that can correctly identify and mark document boundaries.
	
	Also in our experiments, we see that MS-SSIM as a metric does not provide as much attention to line level detail as it does to overall image structure, texture etc. The area dependency of MS-SSIM and LD also causes them to give highly varied results for the same distortion level in images of different areas. Thus, future work on an area independent  standardized metric is highly necessary for proper evaluation of results.



	
	
	%

	\bibliography{main}
	
	
\end{document}